\documentclass{article}

\usepackage[preprint,nonatbib]{neurips_2022}

\usepackage{graphicx}
\usepackage{amsmath}
\usepackage{amssymb}
\usepackage{multirow}
\newcommand{\cmark}{\ding{51}}%
\newcommand{\xmark}{\ding{55}}%

\usepackage{comment}
\usepackage{amssymb}
\usepackage{pifont}
\usepackage{bbm}

\newcommand{\ie}{\textit{i}.\textit{e}.}
\newcommand{\eg}{\textit{e}.\textit{g}.}
\usepackage{wrapfig,lipsum}

\usepackage[utf8]{inputenc} 
\usepackage[T1]{fontenc}    
\usepackage{url}            
\usepackage{booktabs}       
\usepackage{amsfonts}       
\usepackage{nicefrac}       
\usepackage{microtype}      
\usepackage{xcolor}         

\usepackage{multicol}
\usepackage{multirow}
\usepackage{caption}

\usepackage{color,colortbl}
\definecolor{Gray1}{gray}{0.9}
\definecolor{Gray2}{gray}{0.99}

\usepackage[pagebackref,breaklinks,colorlinks]{hyperref}

\title{Normalization Perturbation: \\ 
A Simple Domain Generalization Method \\ for Real-World Domain Shifts}

\author{%
  Qi Fan\thanks{Email: {\tt fanqics@gmail.com}} \\
  HKUST \\
  \And
  Mattia Segu \\
  ETH Zurich \\
  \And
  Yu-Wing Tai \\
  Kuaishou Technology \\
  \And
  Fisher Yu \\
  ETH Zurich \\
  \And
  Chi-Keung Tang \\
  HKUST \\
  \And
  Bernt Schiele \\
  Max Planck Institute for Informatics \\
  \And
  Dengxin Dai \\
  Max Planck Institute for Informatics \\
}

\begin{document}

\maketitle

\begin{abstract}
Improving model's generalizability against domain shifts is crucial, especially for safety-critical applications such as autonomous driving.
Real-world domain styles can vary substantially due to environment changes and sensor noises, but deep models only know the training domain style.
Such domain style gap impedes model generalization on diverse real-world domains.
Our proposed Normalization Perturbation (NP) can effectively overcome this domain style overfitting problem. We observe that this problem is mainly caused by the biased distribution of low-level features learned in shallow CNN layers. Thus, we propose to perturb the channel statistics of source domain features to synthesize various latent styles, so that
the trained deep model can perceive diverse potential domains and generalizes well even without observations of target domain data in training.
We further explore the style-sensitive channels for effective style synthesis.
Normalization Perturbation only relies on a single source domain and is surprisingly effective and extremely easy to implement.
Extensive experiments verify the effectiveness of our method for generalizing models under real-world domain shifts.

\end{abstract}

\section{Introduction}

Deep learning has made great progress on in-domain data~\cite{chen2018encoder,ren2015faster}, but its performance usually degrades under domain shifts~\cite{sakaridis2018semantic,sakaridis2021acdc}, where the testing (target) data differ from the training (source) data.
Real-world domain shifts are usually brought by environment changes, such as different weather and time conditions, 
attributed by diverse contrast, brightness, texture, etc.
Trained models usually overfit to the source domain style and generalize poorly in other domains, posing serious problems in 
challenging real-world usage such as autonomous driving. 
Domain generalization (DG)~\cite{muandet2013domain,ghifary2016scatter,mahajan2021domain,li2020domain}, as well as  unsupervised domain adaptation (UDA) methods~\cite{schneider2020improving,nado2020evaluating,hoyer2022daformer}, aims to solve this hard and significant problem.

\begin{figure}
    \centering
    \includegraphics[width=0.92\linewidth]{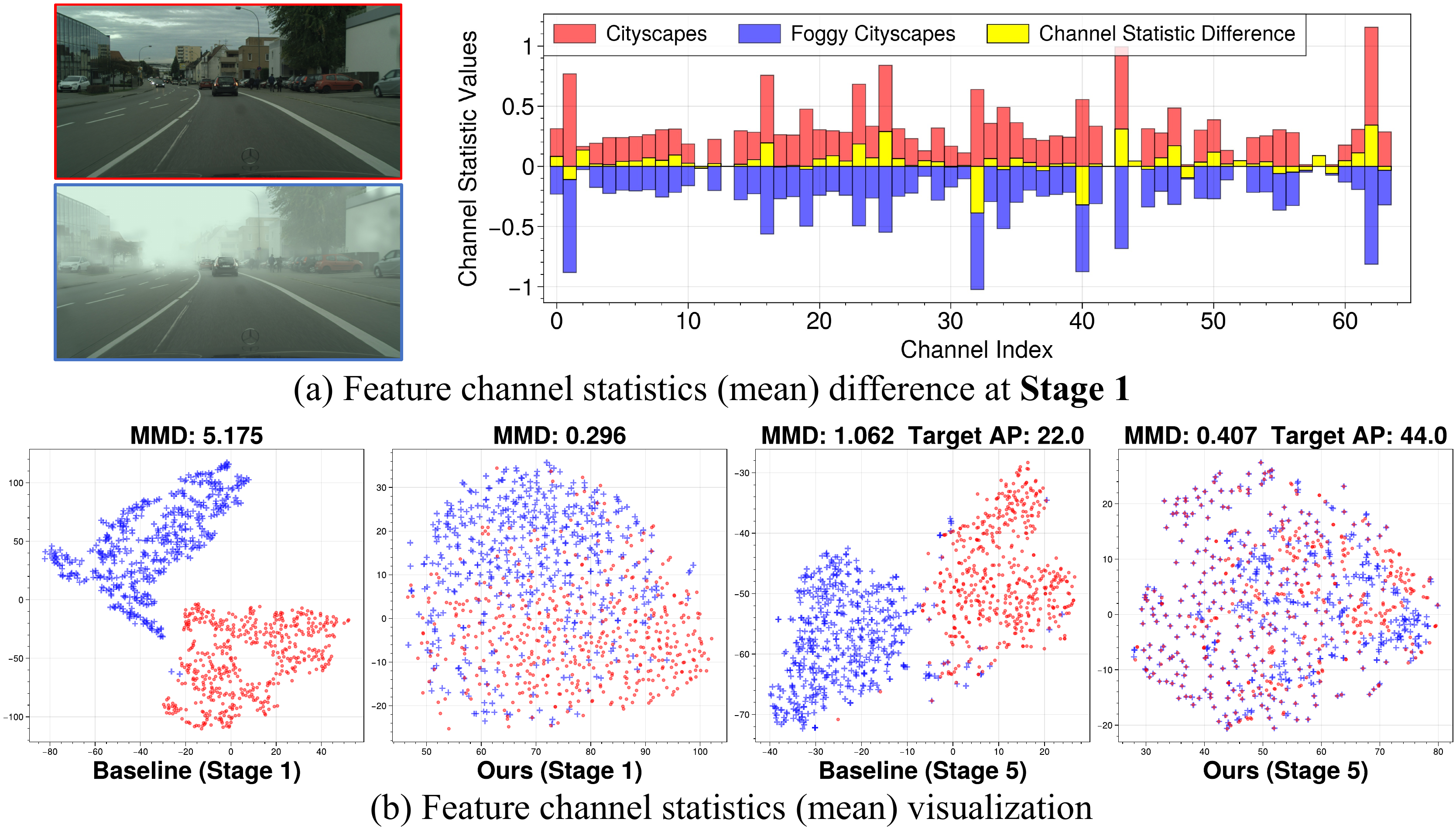}
    \caption{Visualizations for feature channel statistics on \textcolor{red}{Cityscapes (source domain, red)} and \textcolor{blue}{Foggy Cityscapes (target domain, blue)}.
    (a) For two domain images with the same content but different styles, we show their feature channel statistics and differences on the pretrained backbone at stage 1. The statistic values of the Foggy Cityscapes image are converted to negative for better visualization.
    The feature channel statistics of the target domain image deviate around the source domain statistics.
    (b) The t-SNE~\cite{van2008visualizing} visualization for the feature channel statistics on different stages. The model is trained on the source domain and evaluated on both domains.
    The 
    distance between two domains is computed by Maximum Mean Discrepancy~\cite{borgwardt2006integrating} (MMD).
    After equipping Normalization Perturbation (NP) in shallow CNN layers, our model can effectively blend distinct domain style distributions. The target domain distribution can be properly covered by the perturbed source domain distribution in the deep CNN layers. Thus our model generalizes much better on the target domain.
    }
    \label{fig:teaser}
    \vspace{-0.13in}
\end{figure}

Deep models do not generalize well to unseen domains because they only know the training domain style.
Figure~\ref{fig:teaser}(b) shows a large gap of feature channel statistics between two distinct domains: Cityscapes~\cite{cordts2016cityscapes} and Foggy Cityscapes~\cite{sakaridis2018semantic}, especially in shallow CNN layers which preserve more style information.
Deep models trained on the source domain cannot generalize well on the target domain, due to the discrepancy in feature channel statistics caused by the domain style overfitting.

Ideally, if a model can perceive a large variety of potential domains during training, it can learn domain-invariant representations and generalizes well.
However, it is expensive and even impossible to collect data for all possible domains.
Synthesizing new domains is a feasible solution.
But existing domain synthesis methods still require diverse style sources and can only generalize well to limited domain styles.
The image generation based synthesis approach~\cite{geirhos2018imagenet,carlucci2019hallucinating,gong2019dlow} is powerful but inefficient.
The feature-level synthesis approach~\cite{zhou2020domain,jin2021style,mancini2020towards} is efficient but relies on multiple source domains and the synthesized styles are limited.
In this paper, we propose a novel domain style synthesis approach with high efficiency and effectiveness for real-world DG.

Figure~\ref{fig:teaser}(a) shows our motivation: feature channel statistics of the target domain image deviate around the source domain statistics.
Thus by perturbing the feature channel statistics of source domain images in the shallow CNN layers, we can effectively synthesize new domains.
The perturbed feature statistics correspond to various latent domain styles, so that the trained model perceives diverse potential domains accordingly.
Such perturbation  enables deep models to learn domain-invariant representations where distinct domains can be effectively blended together in the learned feature space. To further boost the performance, we have also explored the style-sensitive channels for effective domain style synthesis. Figure~\ref{fig:teaser}(b)\footnote{For all t-SNE visualizations in this paper, the features from multiple models are mapped jointly into a unified space but are separately visualized for clarity.} shows the distinct domains can be effectively blended by the perturbed channel statistics in shallow CNN layers.
The learned deep CNN representations are thus more robust to the variations of different domain styles and generalize well on the target domain. Note that the model is only trained using a single source domain data without any access to the target domain style or data.

Our method normalizes features into different scales, thus called as Normalization Perturbation (NP).
Our NP is surprisingly effective and extremely easy to implement.
Our NP generalizes well under various real-world domain shifts and outperforms previous DG and UDA methods on multiple dense prediction tasks.
Besides, our method does not change the model architecture, does not require  any extra input, learnable parameters or loss. In fact, 
nothing needs to be changed, except that we perform feature channel statistics perturbation 
on shallow layers during training for synthesizing diverse potential styles from source domain itself.
In summary, our contributions are:
\vspace{-0.1in}
\begin{itemize}
\setlength\itemsep{-0.2em}
    \item We investigate the real-world domain shifts and observe that the domain overfitting problem is mainly derived from the biased feature distribution in low-level layers.
    \item We propose to perturb the channel statistics of source domain features to synthesize various new domain styles, which enables a model to learn domain-variant representations for good domain generalization.
    \item Our method generalizes surprisingly well under various real-world domain shifts and is extremely easy to implement, without any extra input, learnable parameters or loss. Extensive experiments verify the effectiveness of our method.
\end{itemize}

\section{Related Works}

Domain Generalization (DG)~\cite{hendrycks2018benchmarking,saenko2010adapting,li2017deeper,wang2018learning,fang2013unbiased,venkateswara2017deep}, which targets at generalizing models to unseen domains, relies on source data typically consisting of multiple distinct domains. 
DG has been mainly studied in the context of object recognition task~\cite{motiian2017unified,balaji2018metareg,li2019feature,li2019episodic,zhao2021learning,du2020learning,du2020metanorm,yue2019domain}. DG of dense prediction tasks~\cite{qiao2020learning,volpi2019addressing,choi2021robustnet,seemakurthy2022domain,lin2021domain,zhang2022towards,michaelis2019benchmarking,wang2021robust} has attracted increasing interests because of its wide real-world applications.
The closely related unsupervised domain adaptation (UDA)~\cite{ganin2015unsupervised,ghifary2016deep,bousmalis2016domain,bousmalis2017unsupervised,ganin2016domain,long2016unsupervised,damodaran2018deepjdot} has been widely studied on dense prediction tasks~\cite{vs2021mega,2021Seeking,d2020one,chen2018domain,inoue2018cross,kim2019diversify,saito2019strong,zhu2019adapting,chen2017no,zhang2017curriculum,sankaranarayanan2018learning,zhang2018fully,wu2018dcan,zou2018unsupervised} for real-world applications, which aims to generalize model to the target domain by accessing its unlabeled images.
Both DG and UDA share significant overlap in technicalities, such as domain alignment~\cite{li2018domain,jia2020single,zhao2020domain,peng2019moment,zhou2021domain,wei2021toalign,wei2021metaalign,chen2019progressive,kang2018deep,sun2016deep,kumar2018co,morerio2018minimal}, self-supervised learning~\cite{carlucci2019domain,bucci2021self,ghifary2015domain,wang2020learning,xie2020self,yue2021prototypical,iqbal2020mlsl}, feature disentanglement~\cite{li2017deeper,khosla2012undoing,chattopadhyay2020learning,piratla2020efficient,wang2020cross,cai2019learning,liu2018detach,yang2019unsupervised}, and data augmentation~\cite{shi2020towards,xu2020robust,somavarapu2020frustratingly,borlino2021rethinking,zhou2021semi,choi2019self,wang2021consistency,perone2019unsupervised,wang2022continual}. Our method closely relates to the following works.

{\bf Normalization-based Methods.}
The normalization layers are leveraged to improve model generalization ability.
Various normalization variants have been proposed, such as domain-specific BatchNorm~\cite{liu2020ms,seo2020learning,mancini2018robust,chang2019domain}, AdaBN~\cite{li2016revisiting}, PreciseBN~\cite{wu2021rethinking}, Instance-Batch Normalization~\cite{choi2021meta,pan2018two}, Adversarially Adaptive Normalization~\cite{fan2021adversarially}, Switchable Normalization~\cite{luo2019switchable}, and Semantic Aware Normalization~\cite{peng2022semantic}. 
The test-time adaptation~\cite{wang2020tent,kundu2020universal,zhang2021memo,iwasawa2021test,sun2020test,sun2017correlation,sun2017correlation,gretton2009covariate} attempts to estimate accurate normalization statistics for the target domain during testing.
These methods fit normalization layers to the specific target domains,
while our method normalizes features into different scales to implicitly synthesize arbitrary new domains, and is optimization-free.

{\bf Synthesizing New Domains.}
Data augmentation~\cite{volpi2019addressing,shi2020towards,otalora2019staining,chen2020improving,zhang2020generalizing} has been widely used to synthesize new domains in DG and UDA.
Some methods synthesize new domain images using image-to-image translation models, such as the random~\cite{xu2020robust} or learnable augmentation networks~\cite{carlucci2019hallucinating,zhou2020learning,zhou2020deep,wang2021learning}, and style transfer models~\cite{yue2019domain,zhou2021semi,somavarapu2020frustratingly,borlino2021rethinking}.
Other works propose to perform implicit domain synthesis through the feature-level augmentation~\cite{zhou2020domain,jin2021style,mancini2020towards,tang2021crossnorm,nuriel2021permuted,gong2019dlow} to mix CNN feature statistics of distinct domains to significantly improve the domain synthesis efficiency.

The above methods rely on image generation or multiple source domains for new domain synthesis, and thus the efficiency or effectiveness is limited.
On the other hand, our method only relies on a single source domain to diversify by perturbing the feature channel statistics to produce various latent domain styles.

While SFA~\cite{li2021simple} performs the activation-wise feature perturbation, which may destroy meaningful image contents,
our method performs the channel-wise feature statistics perturbation and keeps the image contents unchanged. Note that
A-FAN~\cite{chen2021adversarial} and FSR~\cite{wang2022feature} also perform feature statistics perturbation. But A-FAN~\cite{chen2021adversarial} relies on specially designed adversarial loss, with  the hyperparameters requiring case-by-case tuning.
The FSR~\cite{wang2022feature} needs a learnable network to produce diverse styles.
Their effectiveness is also unknown for dense prediction tasks under real-world domain shifts.
In contrast, our method is extremely simple while surprisingly effective for the real-world applications.


\section{Problem Analysis}
\label{section-3}

We conduct  empirical studies to demonstrate the real-world domain shift problem, so as to motivate our Normalization Perturbation method on the robust object detection task.
The model is trained on the source domain Cityscapes~\cite{cordts2016cityscapes} train set and directly evaluated on Cityscapes and two unseen target domains Foggy Cityscapes~\cite{sakaridis2018semantic} and BDD100k~\cite{yu2020bdd100k} validation sets.
We use the Faster R-CNN~\cite{ren2015faster} model (with ImageNet~\cite{deng2009imagenet}-pretrained ResNet-50~\cite{he2016deep} backbone) as the baseline.
In our analysis experiments, the shallow CNN layers (stage 1 and 2) and all BatchNorm parameters are frozen\footnote{In this case, the shallow CNN layers are biased towards the domain style of the pretrained ImageNet dataset.} by default following the common object detection practice.
Such setting applies to all object detection experiments.
The detection performance is evaluated using the mean average precision (mAP) metric with the threshold of 0.5.
Refer to the supplementary material for full experimental details.

\begin{figure}[t]
  \begin{center}
    \includegraphics[width=0.95\textwidth]{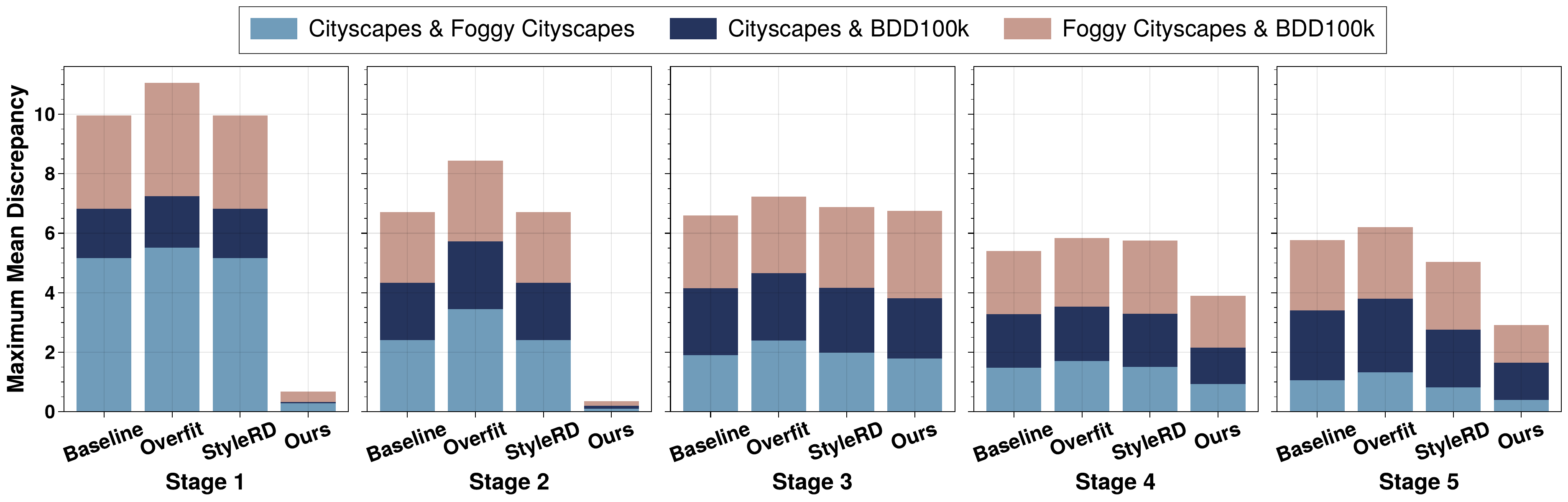}
  \end{center}
  \vspace{-0.1in}
    \caption{Accumulated Maximum Mean Discrepancy (MMD) for the feature channel statistics of different dataset pairs. Four models are evaluated on different convolutional stages. The smaller MMD means smaller feature-level domain/style gap among datasets.}
    \label{fig:mmd}
    \vspace{-0.1in}
\end{figure}

Table~\ref{table-100} shows four Faster R-CNN models used for analysis and their performance on three datasets.

\begin{table}[t] 
  \caption{Four Faster R-CNN models with different settings. They are all trained on Cityscapes train set and evaluated on Cityscapes (C), Foggy Cityscapes (F) and BDD100k (B) val sets.} 
  \vspace{0.05in}
  \centering
      \begin{tabular}{ccccc}
        \toprule
        Method & Description & C & F & B \\
        \midrule
        {\tt Baseline} & The {\it Stage1-2} layers are frozen. & 58.1 & 22.0 & 21.8 \\
        {\tt Overfit} & The {\tt Baseline} with no frozen layers. 
        & {\bf 59.5} & 16.3 & 20.5 \\
        {\tt StyleRD} & The {\tt Baseline} trained with the style randomization.
        & 51.9 & 30.4 & 26.0 \\
        \rowcolor{Gray1} {\tt Ours} & The {\tt Baseline} trained with our Normalization Perturbation. 
        & 58.7 & {\bf 44.0} & {\bf 30.1} \\

        \bottomrule
      \end{tabular}
\vspace{-0.1in}
\label{table-100}
\end{table}

{\bf Biased Model Impedes Domain Generalization.}
With training data under the same domain style, the learned model performs well in testing for data under the same distribution as the training data, with ability of grouping in-domain features together.
But the learned model tends to separate distinct domains and thus hardly generalizes from the source to target domain.
Figure~\ref{fig:teaser} shows image feature channel statistics of the same domain are grouped together, while different domains are separated.
Figure~\ref{fig:mmd} and Table~\ref{table-100} show that the biased distribution in the {\tt Baseline} and {\tt Overfit} models causes large domain feature statistic discrepancy which impedes  model generalization to unseen domains. 

{\bf Shallow CNN Features Matter for Generalization.} 
Figure~\ref{fig:teaser} and Figure~\ref{fig:mmd} show that  shallow CNN layers exhibit larger domain feature statistic discrepancy.
Such discrepancy is propagated from the shallow to deep layers and finally results in the poor target domain performance.
The shallow CNN layers suffer more from severe biased distribution when they are further trained on the source domain.
Note in particular Figure~\ref{fig:mmd} shows that the {\tt Overfit} model has larger domain feature gaps on all layers.
Table~\ref{table-100} further shows quantitatively that this overfitting model generalizes worse on unseen target domains, while capable of producing better source domain performance.
Thus shallow CNN layers do matter for generalizing model to different domain styles, because they preserve more style information through encoding local structures, such as corner, edge, color and texture, which are closely relevant to styles~\cite{zeiler2014visualizing}.
While the deep CNN layers encode more semantic information which are more insensitive to the style effect, if the model is trained on the biased shallow CNN features, the deep layers cannot effectively calibrate the style-biased semantic information and thus the entire model overfits to the source domain.

{\bf Reducing Domain Style Overfitting.}
Diverse training domains would help deep models to learn domain-invariant representations and thus reduce the domain style overfitting.
Our NP efficiently synthesizes diverse latent domain styles and effectively reduces any inherent domain style overfitting.
Figure~\ref{fig:teaser} and Figure~\ref{fig:mmd} show our NP significantly reduces the domain feature gap, especially in the shallow and deep CNN layers. 
Table~\ref{table-100} shows that {\tt Ours} model with NP generalizes well on unseen target domains while simultaneously keeping the source domain performance.
The image-level domain style synthesis method {\tt StyeRD} also reduces domain style gaps and improves domain generalization. However, as we will show shortly, this method is not as desirable as ours.

\section{Method and Analysis}
\label{section-4}

It has been widely studied that feature channel statistics, \eg, mean and standard deviation, tightly relate to image styles.
Changing feature channel statistics can be regarded as implicitly changing the input image styles.
The Adaptive Instance Normalization (AdaIN)~\cite{huang2017arbitrary} achieves arbitrary style transfer through the feature channel statistics normalization and transformation.
Given a mini-batch $B$ of CNN features $x \in \mathcal{R}^{B \times C \times H \times W}$ with $C$ channels and $H \times W$ spatial size from the content images. AdaIN can be formulated as: 
\begin{equation}
    y = \sigma_s \frac{x - \mu_c}{\sigma_c} + \mu_s,
\end{equation}
where both $\{\mu_c, \sigma_c\} \in \mathcal{R}^{B \times C}$ and $\{\mu_s, \sigma_s\} \in \mathcal{R}^{B \times C}$ are feature channel statistics, estimated from the input content images and style images, respectively.
The normalized features $y$ can be decoded into the stylized content images.
AdaIN provides a feasible and efficient way to implicitly change image styles in the feature space.

\begin{figure}[t]
    \centering
    \includegraphics[width=1.0\linewidth]{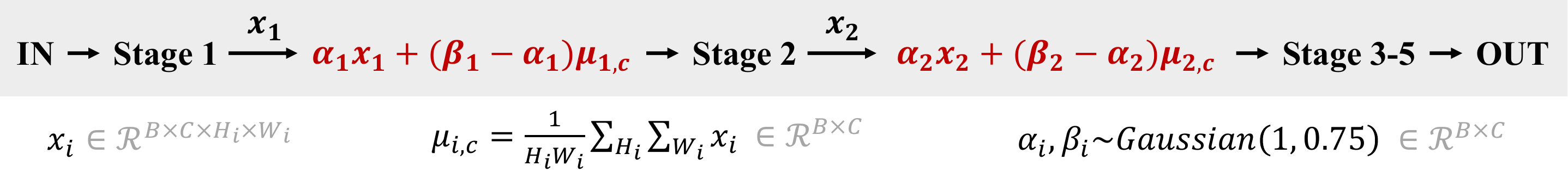}
    \caption{Our Normalization Perturbation (NP) is applied at shallow CNN layers only during training. NP is enabled with probability $p=0.5$.} 
    \label{fig:method}
\end{figure}

\subsection{Normalization Perturbation Method}

Our proposed method, Normalization Perturbation (NP), perturbs the feature channel statistics of training images by inserting random noises.
Formally, NP can be formulated as:
\begin{equation}
    y = \sigma_s^\star \frac{x - \mu_c}{\sigma_c} + \mu_s^\star, \qquad
    \sigma_s^\star = \alpha \sigma_c, \qquad
    \mu_s^\star = \beta \mu_c
\end{equation}
where $\{\mu_c, \sigma_c\} \in \mathcal{R}^{B \times C}$ are the channel statistics, mean and variance, estimated on the input features. The
$\{\alpha, \beta\} \in \mathcal{R}^{B \times C}$ are random noises drawn from the Gaussian distribution.
This equation can be further simplified as:
\begin{equation}
    y = \alpha x + (\beta - \alpha) \mu_c.
\end{equation}
As shown in Figure~\ref{fig:method}, NP is applied at shallow CNN layers (following the backbone stage 1 and stage 2). 
NP is enabled with probability $p$ only in the training stage.

Our NP method is fundamentally different from conventional normalization methods~\cite{huang2017arbitrary,ioffe2015batch,ulyanov2016instance,wu2018group}, whose affine parameters $\{\mu_s, \sigma_s\}$ are learned from the training set or estimated from extra input style images. 
While NP affine parameters $\{\mu_s^\star, \sigma_s^\star\}$ are obtained by perturbing the input feature channel statistics,
they are obtained without relying on extra style inputs and are optimization-free.
The perturbed affine parameters can be regarded as the channel statistics corresponding to diverse latent domain styles, 
enabling models to learn domain-invariant representations and preventing them from style overfitting.

In NP, all channel statistics are randomly perturbed with the same noise distribution.
We further propose Normalization Perturbation Plus (NP+) to adaptively control the noise magnitude in different channels, based on the feature statistic variance across different images.
Such adaptive perturbation is motivated by the observation that some channels significantly vary  as the domain style changes.
We thus apply more noise on these style-sensitive channels.
Specifically, we use the mini-batch of $B$ feature channel statistics $\mu_c = \{\mu_c^1, ..., \mu_c^b, ..., \mu_c^B\}$ to compute the statistic variance $\Delta \in \mathcal{R}^{1 \times C}$:
\begin{equation}
    \Delta = \frac{1}{B} \sum_{b=1}^B (\mu_c^b - \bar \mu_c) ^ 2, \qquad \bar \mu_c = \frac{1}{B} \sum_{b=1}^B (\mu_c^b).
\end{equation}
Then we use the normalized statistic variance $\delta = \Delta / {\tt max} (\Delta) \in \mathcal{R}^{1 \times C}$ to control the injected noise magnitude for each channel:
\begin{equation}
    y = \alpha x + \delta (\beta - \alpha) \mu_c,
\end{equation}
where {\tt max} is the maximum operation.
When applying NP+, we use the photometric data augmentation (Color Jittering, GrayScale, Gaussian Blur and Solarize, only for NP+ by default) to generate pseudo domain styles to facilitate the exploration on style-sensitive channels.

\subsection{Normalization Perturbation Advantages}

\begin{figure} 
  \begin{center}
    \includegraphics[width=1.0\textwidth]{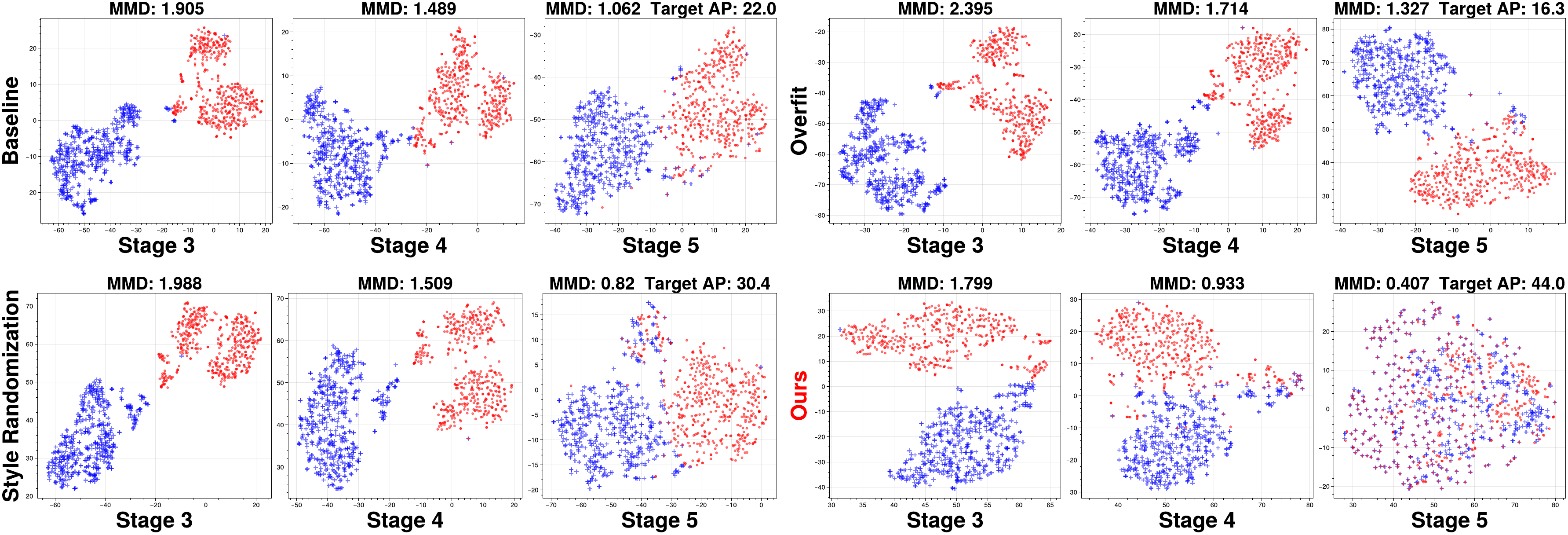}
  \end{center}
    \vspace{-0.1in}
    \caption{The t-SNE visualization for the feature channel statistics of different methods on \textcolor{red}{Cityscapes (source domain)} and \textcolor{blue}{Foggy Cityscapes (target domain)}. The target domain performance is presented.}
    \label{fig:analysis}
\end{figure}

\begin{table}[t]
\caption{Ablation studies on Normalization Perturbation (NP). SR denotes style randomization.}
\vspace{0.05in}
  \footnotesize
  \renewcommand{\tabcolsep}{1.0mm}
\begin{minipage}[t]{0.35\linewidth}
\centering
  \begin{tabular}[t]{cccc}
    \toprule
    Method & C & F & B \\
    \midrule
    Baseline & 58.0 & 22.0 & 21.8 \\
    Image-level SR & 51.9 & 30.4 & 26.0 \\
    Feat-level SR & 58.2 & 42.0 & 29.0 \\
    \rowcolor{Gray1} Ours & 58.7 & 44.0 & 30.1 \\
    \bottomrule
  \end{tabular}
\\
\vspace{0.05in}
(a) Effect of feature-level latent styles.
\end{minipage}
\begin{minipage}[t]{0.32\linewidth}
\centering
  \begin{tabular}[t]{cccc}
    \toprule
     & C & F & B \\
    \midrule
    B(0.75, 0.75) & {\bf 59.0} & 43.0 & 29.5 \\
    U(0, 2.0) & 58.4 & 42.0 & 28.9 \\
    G(1, 0.50) & 58.3 & 40.1 & 29.6 \\
    \rowcolor{Gray1} G(1, 0.75) & 58.7 & 44.0 & 30.1 \\
    G(1, 1.00) & 57.4 & {\bf 44.3} & {\bf 30.2} \\
    \bottomrule
  \end{tabular} 
\\
\vspace{0.05in}
(b) Effect of noise types.
\end{minipage}
\begin{minipage}[t]{0.32\linewidth}
\centering
  \begin{tabular}[t]{cccc}
    \toprule
     & C & F & B \\
    \midrule
    Stage 1 & 59.3 & 40.7 & 29.6 \\
    Stage 2 & 58.4 & 41.5 & 29.5 \\
    Stage 3 & {\bf 59.7} & 27.7 & 24.2 \\
    \rowcolor{Gray1} Stage 12 & 58.7 & {\bf 44.0} & {\bf 30.1} \\
    Stage 123 & 58.3 & 43.8 & 29.9 \\
    \bottomrule
  \end{tabular}
\\
\vspace{0.05in}
(c) Effect of NP positions.
\end{minipage}
\label{table-101}
\end{table}

Our Normalization Perturbation can be implemented as a plug-and-play component in modern CNN models to effectively solve the domain style overfitting problem.
NP has multiple advantages.

{\bf Effective Domain Blending.} Our NP can effectively blend feature channel statistics of different domains, corresponding to learning better domain-invariant representations. 
Figure~\ref{fig:analysis} shows that the {\tt Ours} model trained with NP can effectively reduce the learned distribution distance between source and target domains, especially on deep CNN layers.
Compared to other methods, NP results in smaller cross-domain distribution distance and better generalization performance on target domains.

{\bf High Content Fidelity.}
Our NP processes feature channel statistics while keeping  image and feature spatial structures unchanged.
Note that image-level domain synthesis methods may  destroy the content structures of the original images in the image generation procedure.
Besides, NP trains deep models with numerous content-style combinations in the high-dimensional feature space, which is much more efficient and effective than the image-level methods, whose styles are deterministic and limited and the style augmentation is only performed on the low-dimensional image space.

Table~\ref{table-101}(a) shows the comparisons between image- and feature-level domain synthesis methods.
Image-level style randomization~\cite{geirhos2018imagenet} sacrifices the source domain performance due to its potential destruction on image contents, although this method effectively improves the detection performance on unseen target domains.
We further compare this method with  feature-level style randomization, which is similar to our NP method except that its affine parameters are obtained from extra input style images.
The feature-level style randomization performs well on both source and target domains, while
our method performs best on all datasets due to our diverse latent styles, even without extra style information.

\begin{figure}
  \begin{center}
    \includegraphics[width=0.9\textwidth]{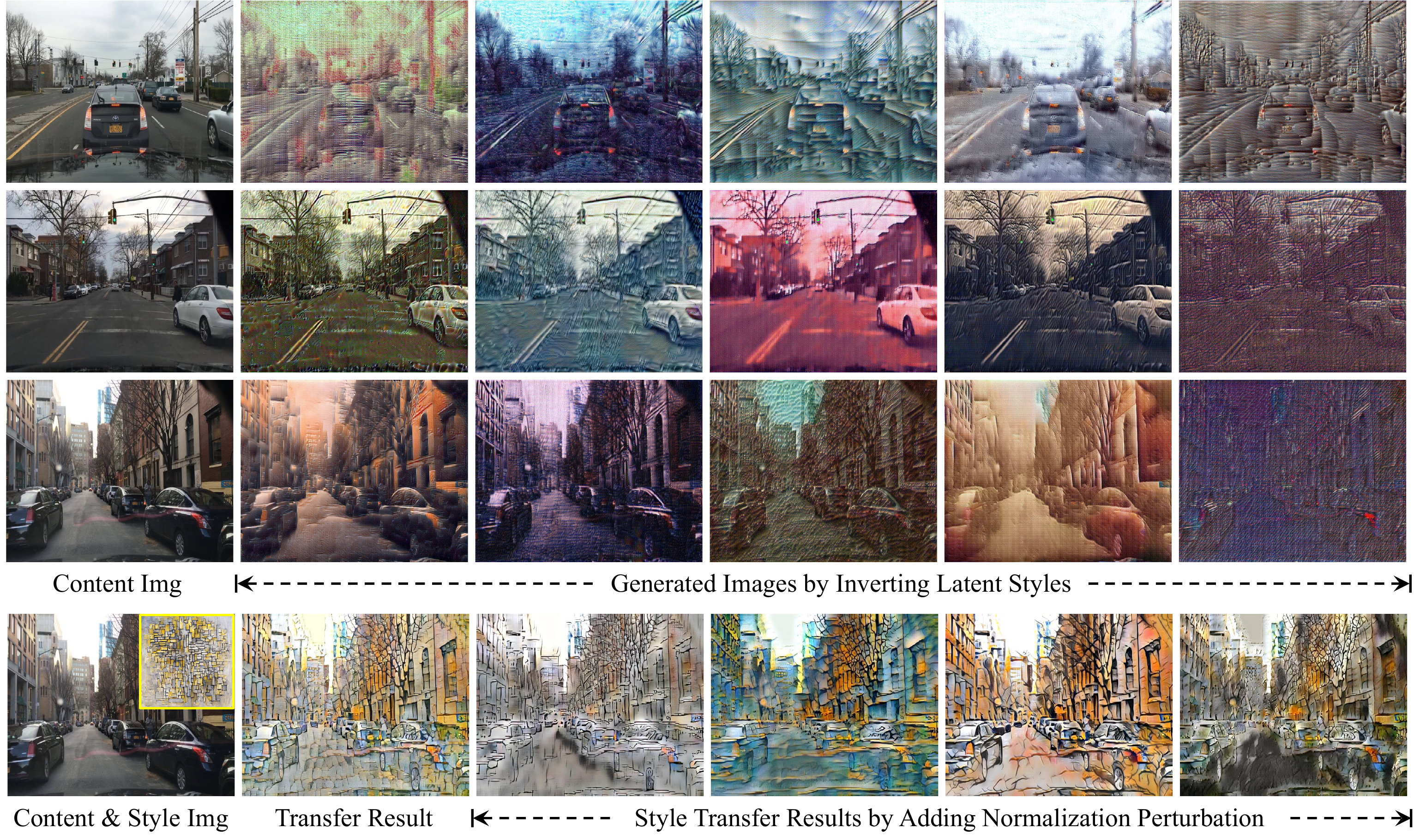}
  \end{center}
  \vspace{-0.05in}
    \caption{Latent style visualization. Top: the perturbed channel statistics of the training image features are inverted back to the image space. Bottom: style transfer results of adding our perturbation.}
    \label{fig:vis}
\vspace{-0.1in}
\end{figure}

{\bf Diverse Latent Styles.}
Our NP effectively diversify the  
latent styles.
For better understanding, we map the NP perturbed feature channel statistics back into the image space using the feature inverting technique~\cite{wang2021rethinking,zeiler2014visualizing}. Figure~\ref{fig:vis} shows that the generated latent styles are diverse to effectively enlarge the style scopes, covering various potential unseen domain styles in real-world environments, \eg, dawn, dusk, night times and foggy, rainy, snowy weathers.


\begin{table}[t]
\caption{Ablation studies on Normalization Perturbation (NP). DA denotes data augmentation.}
\vspace{0.05in}
  \footnotesize
  \renewcommand{\tabcolsep}{1.2mm}
\begin{minipage}[t]{0.33\linewidth}
\centering
  \begin{tabular}[t]{cccc}
    \toprule
     & C & F & B \\
    \midrule
    Baseline & 58.1 & 22.0 & 21.8 \\
    Spatial & {\bf 59.6} & 24.9 & 22.6 \\
    Activation & 58.1 & 25.3 & 26.4 \\
    \rowcolor{Gray1} Channel & 58.7 & {\bf 44.0} & {\bf 30.1} \\
    \bottomrule
  \end{tabular}
\\
\vspace{0.05in}
(a) Comparison to other noise types.
\end{minipage}
\begin{minipage}[t]{0.34\linewidth}
\centering
  \begin{tabular}[t]{cccc}
    \toprule
     & C & F & B \\
    \midrule
    DA & 57.2 & 35.5 & 30.5 \\
    SN \& CN~\cite{tang2021crossnorm} & 58.1 & 31.7 & 26.6 \\
    MixStyle~\cite{zhou2020domain} & 57.7 & 30.1 & 26.5 \\
    \rowcolor{Gray1} Ours & {\bf 58.7} & {\bf 44.0} & {\bf 30.1} \\
    \bottomrule
  \end{tabular}
\\
\vspace{0.05in}
(b) Comparison to other methods.
\end{minipage}
\begin{minipage}[t]{0.32\linewidth}
\centering
  \begin{tabular}[t]{cccc}
    \toprule
     & C & F & B \\
    \midrule
    NP & 58.7 & 44.0 & 30.1 \\
    NP w/ DA & 57.6 & 45.2 & 32.6 \\
    NP+ w/o DA & {\bf 58.8} & 43.2 & 29.9 \\
    \rowcolor{Gray1} NP+ w/ DA & 58.3 & {\bf 46.3} & {\bf 32.8} \\
    \bottomrule
  \end{tabular}
\\
\vspace{0.05in}
(c) Ablation on NP+.
\end{minipage}
\label{table-102}
\end{table}

\subsection{Normalization Perturbation Ablation Studies}

{\bf Noise types.}
Our NP is insensitive to the noise types and hyperparameters.
Table~\ref{table-101}(b) shows our method works well with Beta, Uniform and Gaussian noises. The only requirement is that the noise should be generated around one to enable the perturbed affine parameters be around the input feature channel statistics.

{\bf NP positions.}
Table~\ref{table-101}(c) 
shows our NP performs best when applied in shallow CNN layers. This is because shallow CNN layers are more style sensitive, which fundamentally affect the entire model training as discussed before.

\begin{figure}[t] 
\begin{minipage}[t]{0.49\linewidth}
  \begin{center}
    \includegraphics[width=0.98\textwidth]{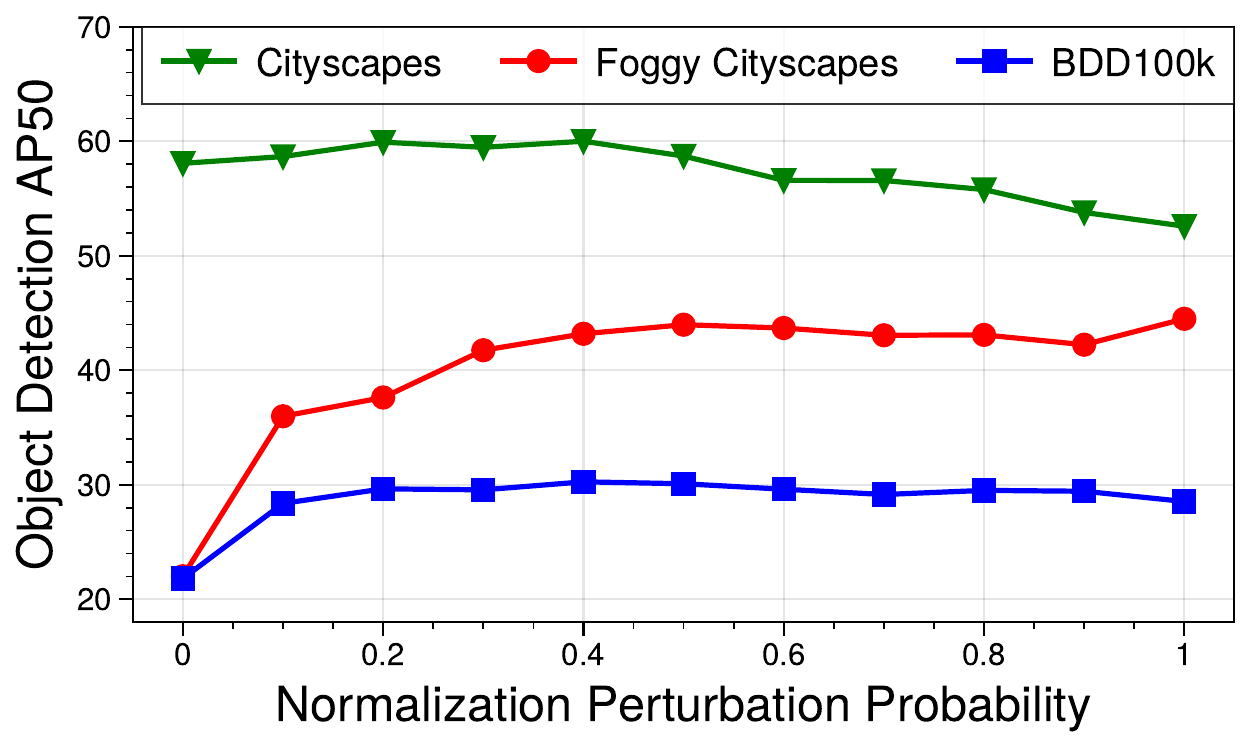}
  \end{center}
    \caption{Effect of NP probability.}
    \label{fig:p}
\end{minipage}
\begin{minipage}[t]{0.49\linewidth}
  \begin{center}
    \includegraphics[width=0.98\textwidth]{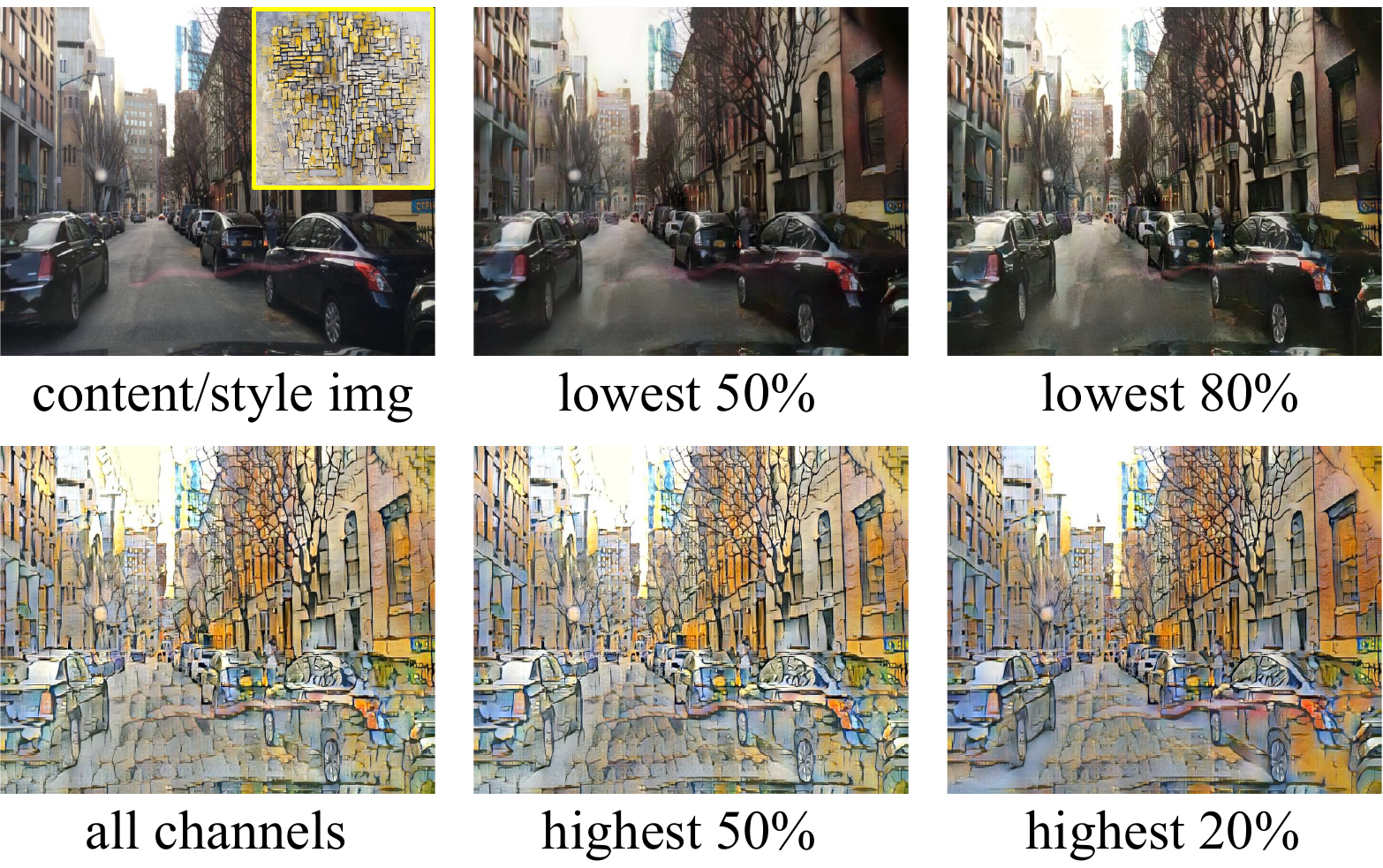}
  \end{center}
    \caption{The style transfer results with style-insensitive/sensitive channels.}
    \label{fig:channel}
\end{minipage}
\end{figure}

{\bf NP 
probability.}
Figure~\ref{fig:p} shows how NP 
probability $p$ affects the model performance.
When the NP 
probability is higher, the generalization performance on target domains tends to be better, but the source domain performance will suffer when the probability is too large.
Interestingly, when NP 
probability is lower than 0.5, NP helps model to perform better on the source domain, probably because NP provides desirable regularization and feature augmentation effect.
We set the probability $p$ as 0.5 in our experiments to achieve the best balance, improving model generalization performance on unseen target domains and simultaneously keeping the source domain performance.

{\bf Perturbation types.} 
NP perturbs feature channel statistics, which is much better than the spatial-level and activation-level~\cite{li2021simple} perturbation as shown in Table~\ref{table-102}(a).
This is because our channel-level perturbation fittingly solves the style overfitting problem, without affecting spatial content structures.

{\bf Comparison to related methods.} 
As shown in Table~\ref{table-102}(b), our method performs better than other methods. 
While photometric data augmentation (DA) improves model generalization, its performance depends on the augmentation and dataset matching degree.
Other feature-level domain synthesis methods~\cite{tang2021crossnorm,zhou2020domain} also work, but are inferior to our method thanks to our diverse latent styles generated by the perturbation operation.

{\bf NP+.}
NP+ adaptively applies heavier perturbation on style-sensitive channels containing larger channel statistic variance.
This is motivated by the observation that style information is mainly preserved at the style-sensitive channels.
As shown in Figure~\ref{fig:channel}, we choose the style-sensitive/insensitive channels to perform style transfer, based on the channel statistic variance between the content and style features.
The style can be effectively transferred into the content image with only the 20\% highest style-sensitive channels, 
while the remaining 80\% channels with low channel statistic variance are style-insensitive and are hardly used to transfer styles.

Table~\ref{table-102}(c) shows the NP+ performs slightly worse than the NP method  without data augmentation because the in-domain images have minor style difference.
But when equipped with photometric data augmentation, the pseudo domain images enable NP+ to effectively find style-relevant channels and thus performs better than NP.

%

\section{Comparison Experiments}

\begin{table}[!t]
  \caption{Robust object detection results.}
  \label{table:bdd100k}
  \centering
  \footnotesize
  \renewcommand{\tabcolsep}{1.0mm}
  \begin{tabular}{c|ccc|ccc|ccc|ccc}
    \toprule
    & \multicolumn{3}{c|}{BDD Day $\to$ Night}  & \multicolumn{3}{c|}{BDD Night $\to$ Day} & \multicolumn{3}{c|}{WaymoL $\to$ BDD} & \multicolumn{3}{c}{WaymoR $\to$ BDD} \\
     & AP & AP50 & AP75 & AP & AP50 & AP75  & AP & AP50 & AP75 & AP & AP50 & AP75 \\

    \midrule
    Faster R-CNN~\cite{ren2015faster} & 17.84 & 31.35 & 17.68 & 19.14 & 33.04 & 19.16 & 10.07 & 19.62 & 9.05 & 8.65 & 17.26 & 7.49 \\
    \midrule
    + Rotation & 18.58 & 32.95 & 18.15 & 19.07 & 33.25 & 18.83 & 11.34 & 23.12 & 9.65 & 9.25 & 18.48 & 8.08  \\
    + Jigsaw & 17.47 & 31.22 & 16.81 & 19.22 & 33.87 & 18.71 & 9.86 & 19.93 & 8.40 & 8.34 & 16.58 & 7.26 \\
    + CycConsist~\cite{wang2021robust} & 18.35 & 32.44 & 18.07 & 18.89 & 33.50 & 18.31 & 11.55 & 23.44 & 10.00 & 9.11 & 17.92 & 7.98  \\
    + CycConf~\cite{wang2021robust} & 19.09 & 33.58 & 19.14 & 19.57 & 34.34 & {\bf 19.26} & 12.27 & 26.01 & 10.24 & 9.99 & 20.58 & 8.30 \\
    \midrule
    \rowcolor{Gray1} + NP (Ours) & 20.73 & 36.22 & 20.85 & 19.32 & 34.42 & 18.63 & 17.85 & 35.34 & 15.52 & 14.97 & 29.42 & 13.11 \\
    \rowcolor{Gray1} + NP+ (Ours) & {\bf 20.97} & {\bf 36.76} & {\bf 21.10} & {\bf 19.73} & {\bf 35.30} & 19.19 & {\bf 21.18} & {\bf 42.16} & {\bf 18.67} & {\bf 19.64} & {\bf 38.69} & {\bf 17.07} \\


    \bottomrule
  \end{tabular}
\label{table-103}
\end{table}

We apply our NP method on two representative real-world dense prediction tasks: object detection and semantic segmentation.
Existing DG and UDA methods mainly focus on one specific task, segmentation or detection.
While our method works on both tasks.


\subsection{Robust Object Detection}

We follow CycConf~\cite{wang2021robust} to train and evaluate models on the robust object detection benchmark.
Specifically, there are two evaluation settings: Domain Shift by Time of Day, where the model is trained on BDD100k~\cite{yu2020bdd100k} daytime/night train set and evaluated on BDD100k night/daytime val set, and Cross-Camera Domain Shift, where the model is trained on Waymo~\cite{sun2020scalability} Front Left/Right train set and evaluated on BDD100k night val set.
Table~\ref{table-103} shows our NP outperforms previous SOTA CycConf~\cite{wang2021robust} on all domain shift settings, especially on the Waymo Front Left/Right to BDD100k Night setting. Our NP+ further boosts the performance to a new SOTA. 

\begin{wraptable}{r}{0.4\linewidth}
\vspace{-0.45in}
  \caption{UDA object  detection AP50 performance on ResNet-50 backbone except $^\dagger$ which are ResNet-101.}
  \vspace{-0.1in}
  \label{table-104-det}
  \centering
  \footnotesize
  \renewcommand{\tabcolsep}{1.0mm}
  \begin{tabular}{cccc}
    \toprule
    Method  & Target &  S $\to$ C& C $\to$ F \\
    \midrule
    FR-CNN~\cite{ren2015faster} & \xmark &  31.9 & 22.8  \\
    DA-Faster~\cite{chen2018domain} & \cmark &  41.9 & 32.0 \\
    DivMatch~\cite{kim2019diversify}  & \cmark & 43.9 & 34.9 \\
    SW-DA~\cite{saito2019strong}  & \cmark &   44.6 & 35.3 \\
    SC-DA~\cite{zhu2019adapting}  & \cmark & 45.1 & 35.9 \\
    MTOR~\cite{cai2019exploring} & \cmark & 46.6 & 35.1 \\
    GPA~\cite{xu2020cross} & \cmark & 47.6 & 39.5 \\
    ViSGA~\cite{2021Seeking}  & \cmark &  49.3 & 43.3 \\
    EPM~\cite{hsu2020every}$^\dagger$  & \cmark &  51.2  & 40.2 \\
    CycConf~\cite{wang2021robust}$^\dagger$   & \cmark &  52.4  & 41.5 \\
    \midrule
    Our Baseline   & \xmark &  32.8 & 22.0 \\
    \rowcolor{Gray1} NP (Ours)    & \xmark & 54.1 & 44.0 \\
    \rowcolor{Gray1} NP+ (Ours)   & \xmark & 58.7 & 46.3 \\
    \bottomrule
  \end{tabular}

\vspace{-0.4in}
\end{wraptable}

\subsection{Unsupervised Domain Adaptive Object Detection}

Unsupervised domain adaptive object detection models are trained on labeled source domain and unlabeled target domain. We consider two popular adaptation settings: Sim10k~\cite{johnson2017driving} to Cityscapes (S $\rightarrow$ C) and Cityscapes~\cite{cordts2016cityscapes} to Foggy Cityscapes~\cite{sakaridis2018semantic} (C $\rightarrow$ F)  adaptations.
Table~\ref{table-104-det} shows that our method significantly outperforms other methods by a large margin {\bf even without accessing the target domain data.}
Our training setting is more practical for real-world applications and surprisingly has much better performance.
Our good performance is properly derived from our improved shallow CNN layers, while other UDA methods attempt to improve deep layers unfortunately based on the biased shallow CNN features.

\subsection{Semantic Segmentation Domain Generalization}

We follow the previous semantic segmentation domain generalization SOTA method RobustNet~\cite{choi2021robustnet} to train and evaluate our method. The model is trained on GTAV/Cityscapes datasets, and evaluated on various datasets, \ie, GTAV (G)~\cite{richter2016playing}, Cityscapes (C)~\cite{cordts2016cityscapes}, BDD100k (B)~\cite{yu2020bdd100k}, Mapillary Vistas (M)~\cite{neuhold2017mapillary}, and Synthia (S)~\cite{ros2016synthia}.
Table~\ref{table-seg} shows that our method performs the best.

\begin{table}
\label{table-105}
  \caption{Semantic segmentation domain generalization results. Train datasets are \underline{underlined}.}
  \centering
  \footnotesize
    \renewcommand{\tabcolsep}{1.3mm}
      \begin{tabular}{ccccccc|cccccc}
        \toprule
        Method & C & B & M & S & \underline{G} & mean & G & B & M & S & \underline{C} & mean \\
        \midrule
        Baseline & 28.95 & 25.14 & 28.18 & 26.23 & 73.45 & 36.39 & 42.55 & 44.96 & 51.68 & 23.29 & {\bf 77.51} & 48.00  \\
        SW~\cite{pan2019switchable} & 29.91 & 27.48 & 29.71 & 27.61 & {\bf 73.50} & 37.64 & 44.87 & 48.49 & 55.82 & 26.10 & 77.30 & 50.52\\
        IBN-Net~\cite{pan2018two} & 33.85 & 32.30 & 37.75 & 27.90 & 72.90 & 40.94 & 45.06 & 48.56 & 57.04 & 26.14 & 76.55 & 50.67 \\
        IterNorm~\cite{huang2019iterative} & 31.81 & 32.70 & 33.88 & 27.07 & 73.19 & 39.73 & 45.73 & 49.23 & 56.26 & 25.98 & 76.02 & 50.64 \\
        ISW~\cite{choi2021robustnet} & 36.58 & 35.20 & {\bf 40.33} & 28.30 & 72.10 & 42.50 & 45.00 & 50.73 & 58.64 & 26.20 & 76.41 & 51.40 \\
        \midrule

        \rowcolor{Gray1} NP (Ours)  & {\bf 40.62} & 35.56 & 38.92 & 27.65 & 72.02 & 42.95 & {\bf 47.87} & 51.09 & 58.60 & 27.02 & 76.24 & 52.16 \\
        \rowcolor{Gray1} NP+ (Ours) & 40.41 & {\bf 36.34} & 38.61 & {\bf 29.25} & 72.27 & {\bf 43.38} & 47.18 & {\bf 52.70} & {\bf 60.13} & {\bf 27.15} & 76.07 & {\bf 52.65} \\
        \bottomrule
      \end{tabular}
\vspace{-0.1in}
\label{table-seg}
\end{table}

\section{Conclusion}
We find that biased shallow CNN layers are one of the main causes of
the domain style overfitting problem under real-world domain shifts. To address the problem, 
we propose Normalization Perturbation (NP) to perturb the channel statistics of source domain features to synthesize various latent styles.
The trained deep model can perceive diverse potential domains and thus generalizes well on unseen domains thanks to the learned domain-invariant representations.
Our NP method only relies on a single source domain to generalize on diverse real-world domains.
Our NP method is surprisingly effective and extremely simple, operates without any extra input, learnable parameters or loss.
Extensive analysis and experiments verify the effectiveness of our Normalization Perturbation.

\bibliographystyle{unsrt}
\bibliography{main}

\end{document}